\pdfoutput=1

\documentclass[11pt]{article}

\usepackage[preprint]{acl}

\usepackage{times}
\usepackage{latexsym}

\usepackage[T1]{fontenc}

\usepackage[utf8]{inputenc}

\usepackage{microtype}

\usepackage{inconsolata}

\usepackage{graphicx}

\title{LLM driven Text-to-Table Generation through Sub-Tasks Guidance and Iterative Refinement}

\author{
 \textbf{Rajmohan C\textsuperscript{1}},
 \textbf{Sarthak Harne\textsuperscript{2}},
 \textbf{Arvind Agarwal\textsuperscript{1}}
\\
 \textsuperscript{1}IBM Research,
 \textsuperscript{2}IIIT-Bangalore
 \\
\texttt{rajmohanc1@in.ibm.com} \\
}

\begin{document}
\maketitle
\begin{abstract}
Transforming unstructured text into structured data is a complex task, requiring semantic understanding, reasoning, and structural comprehension. 
While Large Language Models (LLMs) offer potential, they often struggle with handling ambiguous or domain-specific data, maintaining table structure, managing long inputs, and addressing numerical reasoning.
This paper proposes an efficient system for LLM-driven text-to-table generation that leverages novel prompting techniques. Specifically, the system incorporates two key strategies: breaking down the text-to-table task into manageable, guided sub-tasks and refining the generated tables through iterative self-feedback.
We show that this custom task decomposition allows the model to address the problem in a stepwise manner and improves the quality of the generated table. Furthermore, we discuss the benefits and potential risks associated with iterative self-feedback on the generated tables while highlighting the trade-offs between enhanced performance and computational cost.
Our methods achieve strong results compared to baselines on two complex text-to-table generation datasets available in the public domain.

\end{abstract}

\section{Introduction}
\begin{figure}[t]
  \includegraphics[width=\columnwidth]{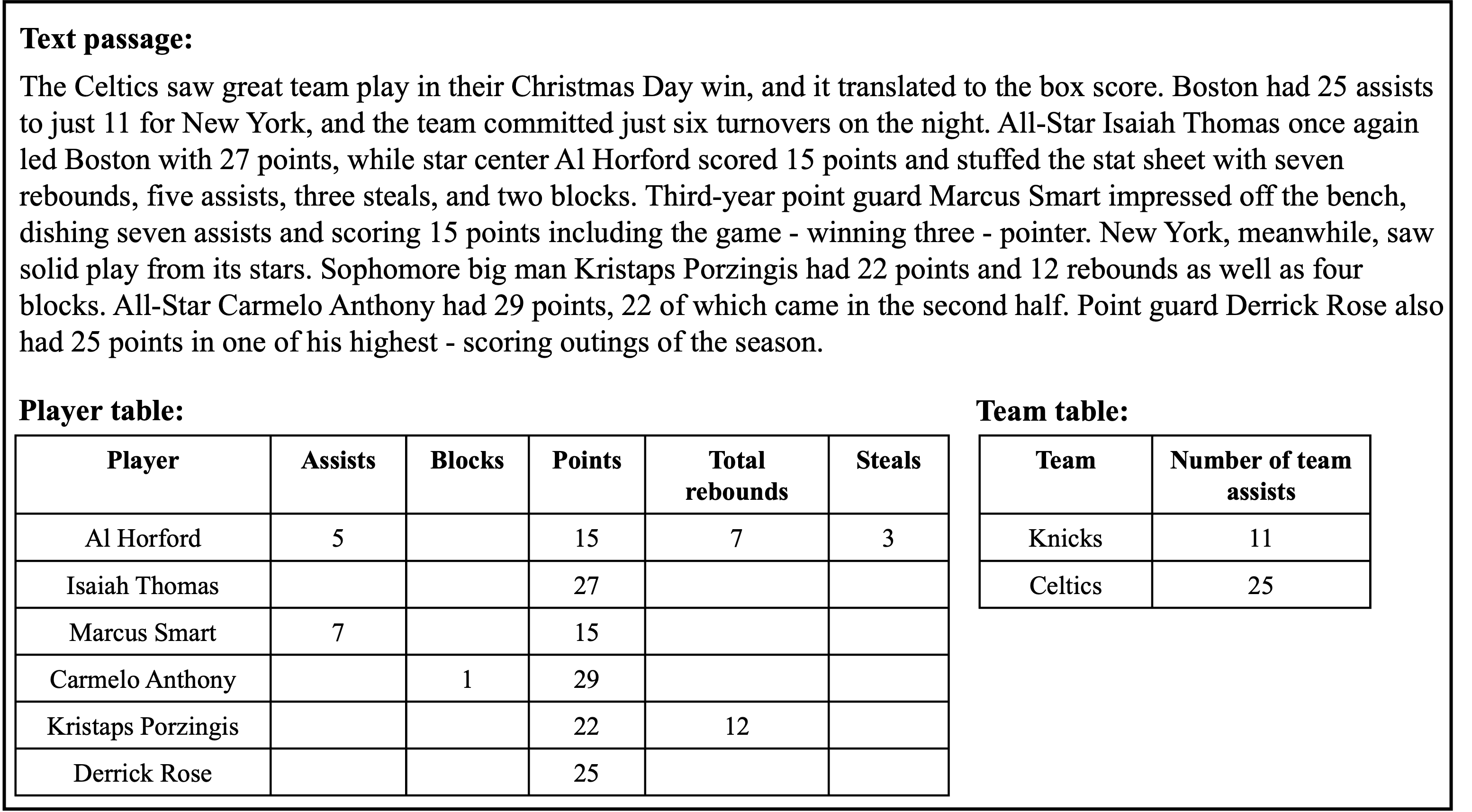}
  \caption{An example of a text passage and its corresponding tables from the \texttt{Rotowire} dataset.}
  \label{fig:rotowire-example}
\end{figure}

The text-to-table generation task transforms a large text passage into a structured table that both accurately represents the facts and faithfully adheres to the information provided in the text. This task has numerous real-world applications including Summarization, Data integration, Knowledge graph construction, Question answering and Analytics \cite{text-to-table}. Figure \ref{fig:rotowire-example} illustrates a text-to-table example, where the input is a basketball game report in text, and the output consists of two tables capturing the scores of players and teams\cite{rotowire}. 
Automatic text-to-table generation is a challenging task, especially when dealing with large, unstructured text that involves several entities, and requires generating tables with numerous rows and columns.
It demands a deep semantic understanding of unstructured text, ability to reason about various entities and their interrelationships and finally align and organize them within tables. 

Large Language Models (LLMs) have demonstrated remarkable capabilities in addressing various natural language processing tasks\cite{achiam2023gpt, anthropic2024introducing, dubey2024llama}.
Tables are well-studied in LLM literature but vast majority of the work focused on tasks where tables are inputs\cite{fang2024large, lu2024large}. However, their application in generating complex structured data as output from a given text remains relatively underexplored. 
Most previous research has concentrated on areas such as entity and event extraction, as well as key-value pairs and relation extraction from text\cite{xu2023large, pai2024survey, ni2023unified}. Some recent studies have begun to explore the use of LLMs for generating structured data \cite{strucbench, gtbls, texttupletable}, but several significant challenges persist in this area of research. 
Smaller models often fall short in handling a complex task like text-to-table generation when used off-the-shelf, due to their limited capacity to follow intricate instructions, manage extensive contexts and apply domain-specific knowledge accurately. 
But fine-tuning these models for this specific task is resource-intensive and often impractical.

In this paper, we focus on LLM prompting based strategies for text-to-table generation, specifically targeting entity-centric tables where each row represents a unique entity (e.g., a player, team, or product), and each column captures attributes or properties of that entity. We start with outlining some of the practical challenges based on our exploratory experiments. We then propose a system addressing some of these challenges via two key components: (1) Explicitly guiding LLMs via custom sub-tasks and (2) Refining table output by leveraging self-feedback of LLMs iteratively. Finally, we present our experimental setup and evaluate the efficiency and performance of the proposed methods compared to existing approaches. We show that our system achieves state-of-the-art performance on two major text-to-table datasets from the public domain. 
To the best of our knowledge, this is the first system to apply custom sub-tasks guidance and iterative self-refinement strategies for the text-to-table generation task.

\begin{figure*}[t]
\centering
    \includegraphics[width=\linewidth]{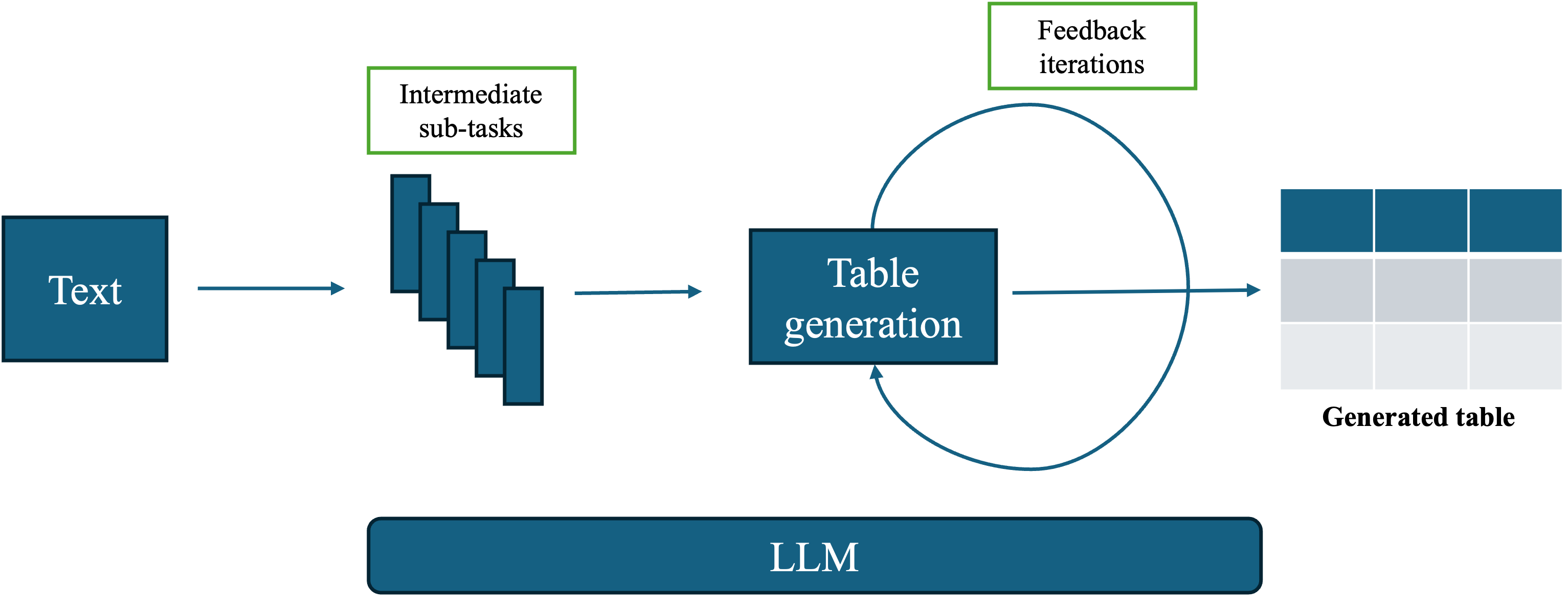}
\caption{LLM driven Text-to-Table Generation via Intermediate sub-tasks guidance \& Output refinement}
\label{fig:sysdiagram}
\end{figure*}

\subsection{Practical Challenges}
\noindent \textbf{Incorrect/missing data:}
The LLM may either generate incorrect data for a table cell or fail to populate a cell altogether based on the input passage. 
This issue arises due to several factors: (1) Entity Resolution—LLMs often struggle to accurately assign data to the correct entity, especially when passages discuss multiple entities simultaneously or contain ambiguous references like pronouns; (2) Domain-Specific Abbreviations—while LLMs can interpret many abbreviations using prior knowledge, they sometimes fail to handle specific ones, leading to missing data; and (3) Data Format Resolution—custom data formats, such as hyphenated scores representing different metrics 
can result in misinterpretation and incorrect table entries.

\noindent \textbf{Table format generation issues:} LLMs sometimes struggle with understanding and maintaining the structure of the generated table. For instance, they may produce tables with inconsistent row lengths or varying numbers of columns. This issue is more prevalent in smaller LLMs, while larger ones generally handle table formatting more reliably, though performance varies by format and model.

\noindent \textbf{Long inputs and few-shot experiments:}
LLMs have limited context windows, which pose challenges for large input passages and tables, especially in token-heavy formats like JSON. It becomes more problematic in few-shot experiments as they require multiple examples within the same context window.

\noindent \textbf{Evaluation issues:}
Automated metrics based on exact or semantic matching often fall short in evaluating text-to-table generation, as tables contain both text and numeric data with distinct evaluation needs. Moreover, assessing the structure of the generated table adds further complexity.

Other significant challenges include the cost of using larger models, as it can quickly escalate, especially with repeated model invocations and the scarcity of large, complex text-to-table datasets for model tuning or evaluation.

\section{Related Work}
\subsection{Text-to-Table Generation} In their pioneering work, \cite{text-to-table} formalized the text-to-table problem as a sequence-to-sequence (seq2seq) task, leveraging transformer-based models to integrate row and column embeddings into the attention mechanism for capturing relationships between headers and data cells better. Building on this, \cite{li2023sequence} introduced a hybrid model that combines sequence-to-sequence and set-based approaches, addressing challenges in accurately capturing unordered and relational data for table generation.

\subsection{LLM Prompting \& Fine-tuning}
Several recent studies have explored LLM-based approaches for text-to-table generation.
\cite{strucbench} investigates structured table generation by instruction fine-tuning LLMs and analyzing how they perform on various automated metrics and the associated evaluation challenges. But the fine-tuning methods are resource-intensive as they require training the models. Prompting-based methods have also been explored; \cite{gtbls} frames table generation as a cell-wise question-answering task using row and column headers. It has been shown to improve accuracy at the cost of more LLM invocations leading to increased computational overhead. 
Similarly, \cite{texttupletable} uses tuples as an intermediate representation, which are then converted to tables for capturing the key-value relationships.
\cite{structsum} breaks down input text into multiple passages to generate tables, but this approach is less practical when the text is not easily divisible.

\subsection{Decomposing a task \& Output refinement}
\label{subsec:output_refinement}
Decomposing a task into sub-tasks has been explored before in certain settings. \cite{sui2024table} proposed a two-step prompting approach for a downstream table task that first prompts LLM to generate additional knowledge about the table, then using it to frame the second prompt to get the final answer. \cite{ye2023large} maps a complex question into easier sub-questions by decomposing a large table into small tables. On LLM self-refinement, \cite{selfrefine} first proposed that LLM performance can be improved by iteratively refining their outputs and incorporating the feedback along several dimensions on certain tasks. Since then it has been explored in few other contexts \cite{shinn2024reflexion, chen2023teaching}.

\section{Our Methodology}
Figure \ref{fig:sysdiagram} shows the high-level overview of our LLM-driven Text-to-table generation system. The two key components of the system are Intermediate sub-tasks guidance \& Output refinement using iterative self-feedback.

\subsection{Intermediate sub-tasks guidance}
\label{subsec:intermediate_subtasks}
Decomposing a complex task into smaller, manageable sub-tasks is known to mitigate some of the common issues like missing steps or misunderstandings.
Accordingly, we break down the text-to-table generation task into intermediate sub-tasks as shown in Figure \ref{fig:subtasks}. 
This decomposition helps guide Large language models (LLMs) in better understanding both the data and the task, enhancing the clarity in reasoning and ultimately improving performance.
The sub-tasks are as follows:

\begin{figure}[ht]
\centering
    \includegraphics[width=\columnwidth]{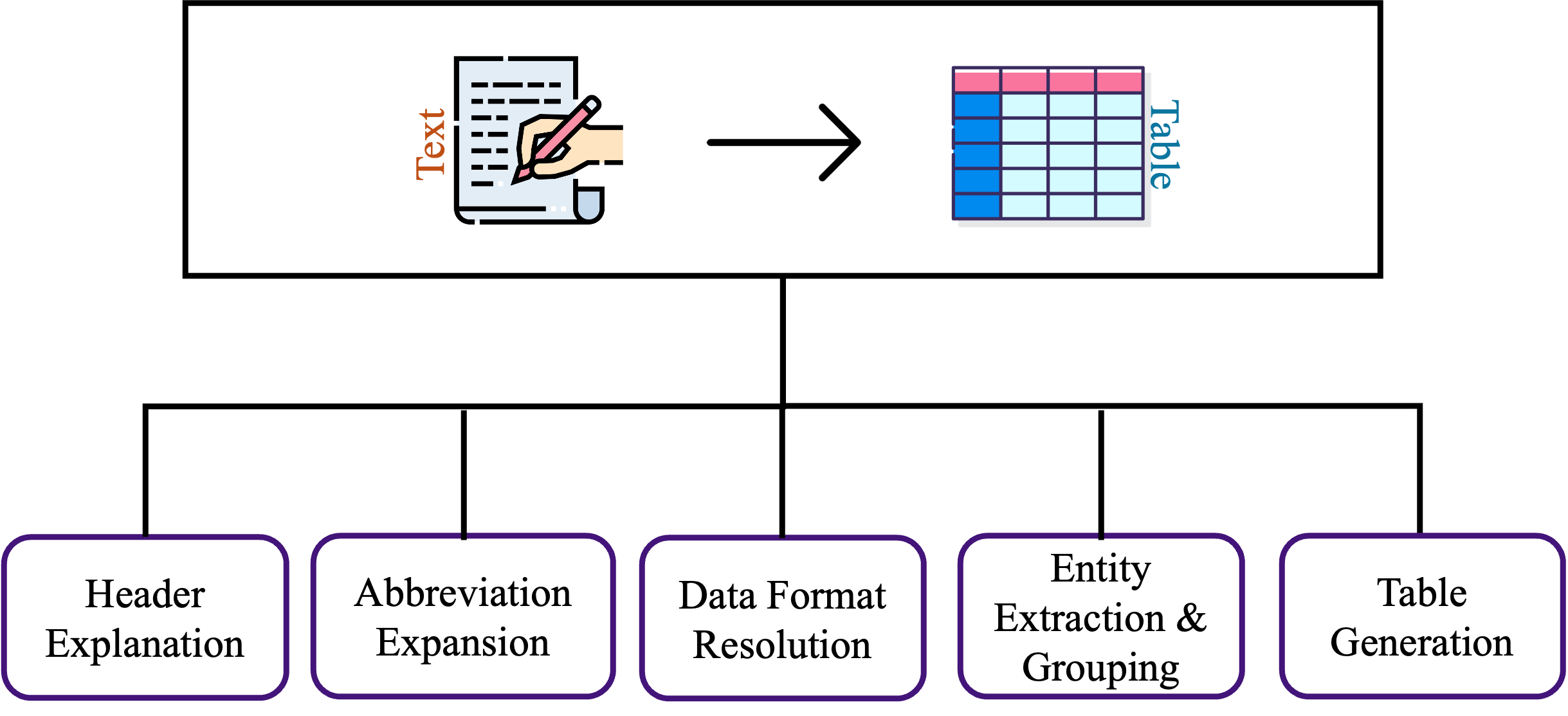}
\caption{Intermediate sub-tasks for text-to-table generation}
\label{fig:subtasks}
\end{figure}

\noindent \textbf{Header Explanation:} In this initial sub-task, the LLM interprets and explains the meaning of the provided column headers in the context of the passage, setting the stage for the table generation.

\noindent \textbf{Abbreviation Expansion:} The LLM expands abbreviations in the passage to ensure full comprehension, reducing the risk of data misinterpretation.

\noindent \textbf{Data Format Resolution:} The LLM resolves custom or non-standard data formats (e.g., hyphenated scores or shorthand notations), to ensure consistency in the final table.

\noindent \textbf{Entity Extraction and Grouping:} The LLM identifies all the row header entities mentioned in the passage and groups data related to each one of them. This process effectively summarizes what will become individual rows in the generated table.
By resolving and organizing text data before the table creation, this step aims to enhance the model's ability to perform accurate entity resolution and association, leading to more coherent and contextually appropriate table rows.

\noindent \textbf{Table Generation: } In this final sub-task, the LLM generates the table in a structured format(slightly similar to the one used in \cite{xie2022unifiedskg, liu2021tapex}), where rows are separated by new lines and cells by the pipe symbol ($|$) , ensuring the output is both machine-readable and human-interpretable.

\subsection{Self-Refinement using Feedback}
\label{subsec:self_refinement}
\begin{figure}[ht]
\centering
    \includegraphics[width=\columnwidth]{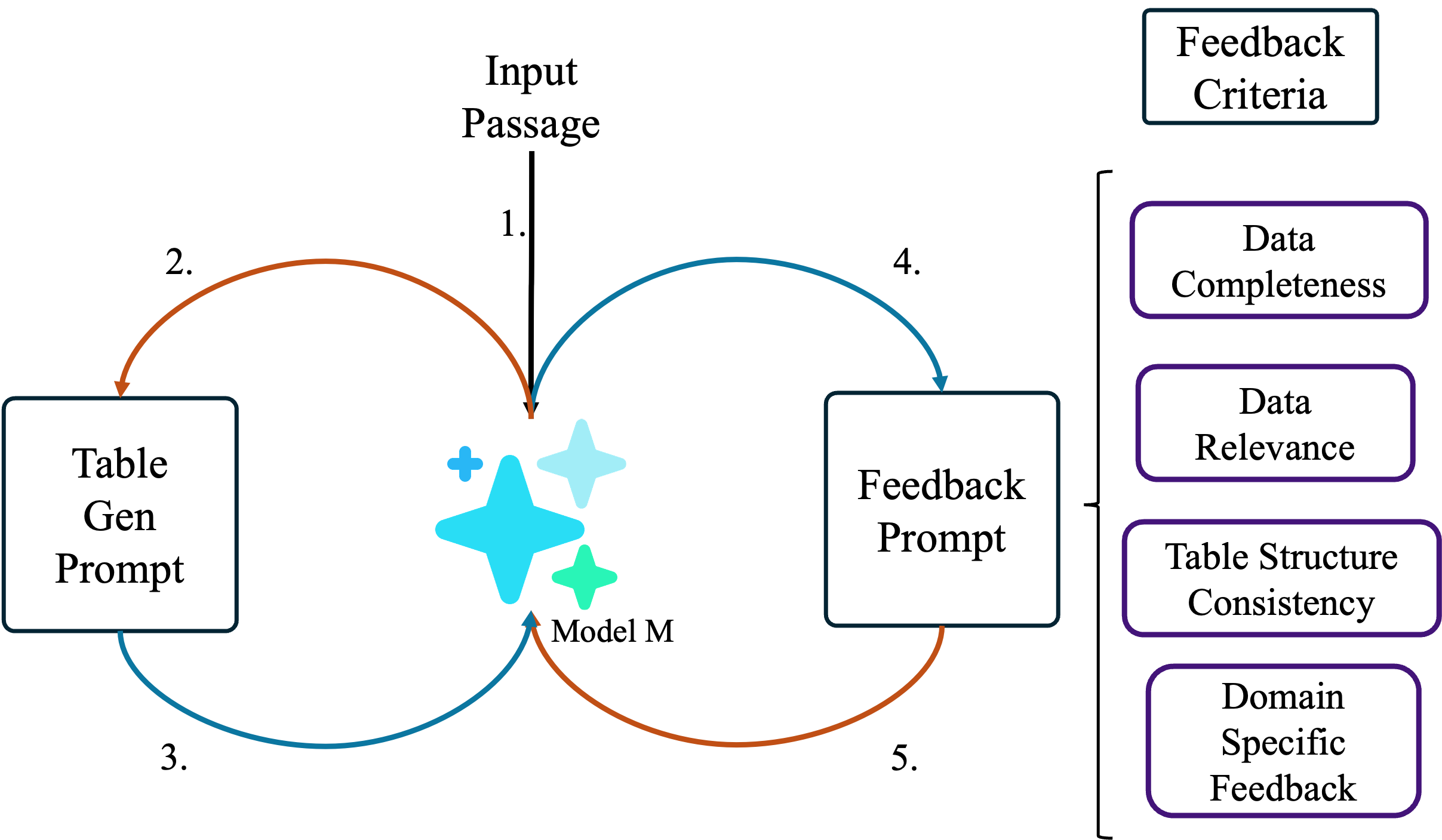}
\caption{Self-refinement for text-to-table generation}
\label{fig:selfrefine}
\end{figure}

We next refine the generated table iteratively using self-feedback of LLMs, to further enhance performance.
After the LLM generates the table, it is prompted to review and provide feedback on its own output as illustrated in Figure \ref{fig:selfrefine}.
The LLM evaluates the generated table using the following criteria: (1) Data Completeness—verifying that all relevant information from the passage is included, addressing any omissions; (2) Data Relevance—ensuring the table contains only information grounded in the passage, avoiding hallucination or extraneous data; (3) Table Structure Consistency—checking that the table adheres to the expected format, including correct row-column alignment and appropriate use of headers; and (4) Domain-Specific Feedback, which involves tailoring the evaluation to specific domain or dataset requirements if any. While this base approach prompts the model to provide feedback on the table as a whole, we also explore granular feedback at the row and cell level to further refine the feedback.

\noindent \textbf{Row-level feedback}: This approach involves asking the model to provide feedback for one row at a time by providing the row header, column headers and generated row data as input.
The rationale behind this is that focusing on individual rows may allow the LLM to examine and identify errors in generation than evaluating the entire table at once.

\noindent \textbf{Cell-level feedback}: This involves asking the LLM to provide feedback for one cell at a time by supplying the specific cell's row header, column header, and generated cell value as input. This method asks the model to verify the accuracy of a particular cell value through recalculation or rechecking. 

\section{Experiments \& Results}

\subsection{Prompting Strategies}
We evaluate our text-to-table generation system, comparing it against baseline methods such as Zero-Shot, Few-Shot, and Chain of Thought (CoT) prompting.
\begin{itemize}
    \item \textbf{Zero-Shot:} The model receives a brief task description, a dataset passage, and the row and column headers for the table. An explanation of the expected table format is also included in the prompt.
    \item \textbf{Few-Shot:} 
    Sample input, output are provided in addition to the above. For cost efficiency, we use a 1-shot example approach.
    \item \textbf{Chain of Thought (CoT):} Here the model is prompted to think step-by-step and solve the task. The steps are determined autonomously by the LLM.
    \item \textbf{Intermediate sub-tasks guidance:} 
    The model is prompted to work through the predefined intermediate sub-tasks(as explained in \ref{subsec:intermediate_subtasks}), designed to guide the generation process before producing the final table.
    
    \item \textbf{Output refinement via feedback:} 
    The LLM evaluates the generated table against predefined criteria (as explained in \ref{subsec:self_refinement}), and regenerates the table after refinement. 

\end{itemize}


\subsection{Models}
We use \textbf{Llama-3-70B-Instruct}, a leading open-license instruction-tuned large language model (LLM), with a token limit of 8k, for our evaluation. 
The model is accessed through the API interface provided by our organization's AI models framework, enabling seamless integration and consistent performance assessment.

\subsection{Datasets}
\noindent \textbf{RotoWire} The RotoWire dataset \cite{rotowire}, initially meant for table-to-text tasks, repurposed as a text-to-table dataset as in \cite{text-to-table}. It contains NBA game summaries paired with two tables: player statistics and team statistics. These tables capture game-related metrics across multiple rows and columns. Given column headers, row headers (player or team names), and an input passage, the primary tasks are extracting relevant information, resolving entities to match data correctly to players/teams, and generating the corresponding tables.

\noindent \textbf{LiveSum} It is a recently published text-to-table dataset featuring football game commentaries \cite{texttupletable}. It contains Team tables with rows for the Home and Away teams and is categorized by difficulty levels based on table columns.  It uses anonymized entities (e.g., PlayerXX, Home/Away Team). It presents greater complexity than the RotoWire dataset as it rarely includes exact field values in the commentary. The main challenge here lies in identifying and counting events like shots, fouls and free kicks.

Unlike simpler datasets like E2E, Wiki-Bio, and WikiTableText \cite{text-to-table}, which primarily involve key-value pair tables, these datasets feature multi-row, multi-column structures that better capture real-world intricacies.

\subsection {Metrics} 
For the Rotowire dataset, we use Exact-Match F1-Score and BERT-Score \cite{bertscore} as evaluation metrics averaged over the Team and Player tables in line with prior research \cite{text-to-table, gtbls, strucbench}. BERT-Score, though less suited for numerical dataset tables, is included for consistency with prior work. We recommend prioritizing Exact Match as the primary evaluation metric.
For the LiveSum dataset, we evaluate using Root Mean Square Error(RMSE) and Cell-level Error Rate(ER), and perform a separate analysis for columns categorized by difficulty levels, as outlined by \cite{texttupletable}.

\subsection{Results}
Table \ref{table:rotowire-our-results} presents the results for the Rotowire dataset, including relevant prior-art (top 4 rows with references, where $^\#$ indicates standard LLM prompting with large models and $^*$ denotes custom prompting strategies that generally involve more LLM invocations). While not directly comparable due to differences in models, their sizes, and prompts, they may still serve as useful reference points.
\setlength{\tabcolsep}{1mm} 
\begin{table}[!ht]
    \centering
        \fontsize{9}{10}\selectfont
        \begin{tabular}{l|cc}
        \hline
            \textbf{Model/Strategy} & \textbf{Exact Match} & \textbf{Bert Score} \\ \hline

            Seq2seq Transformer \\ \cite{text-to-table} & 0.8253 & 0.8897 \\ \hline

            GPT-3.5 Zero-shot$^\#$ & 0.7561 & 0.8374 \\ 
            GPT-3.5 1-Shot$^\#$ & 0.8585 & 0.9723 \\
            \cite{jpjournal}  & & \\ \hline
            ChatGPT Zero-shot$^\#$ & - & 0.9675 \\ 
            GPT-4 Zero-shot$^\#$ & - & 0.9815 \\ 
            ChatGPT Zero-shot + T3$^*$ & - & 0.9734 \\ 
            GPT-4 Zero-shot + T3$^*$ & - & 0.983 \\ 
            \cite{texttupletable} &  & \\ \hline
            Flan-T5-XXL Zero-shot + QA$^*$ \\ \cite{gtbls} & 0.8548 & \textbf{0.9977} \\ \hline \hline
            
            Llama-3-70b-Ins Zero-shot & 0.7053 & 0.8842 \\ \hline
            Llama-3-70b-Ins Zero-shot CoT & 0.6789 & 0.8849 \\ \hline
            Llama-3-70b-Ins 1-Shot & 0.8151 & 0.8989 \\ \hline
            Llama-3-70b-Ins 1-Shot CoT & 0.8433 & 0.9476 \\ \hline
            \begin{tabular}{@{}l@{}} Llama-3-70b-Ins Zero-shot \\ + Intermediate sub-tasks \end{tabular} & 0.8860 & 0.9172 \\ \hline 
            \begin{tabular}{@{}l@{}} Llama-3-70b-Ins Zero-shot \\ + Intermediate sub-tasks \\ + Feedback(Table level) \end{tabular} & 0.8641 & 0.9141 \\ \hline  
            \begin{tabular}{@{}l@{}} Llama-3-70b-Ins Zero-shot \\ + Intermediate sub-tasks \\ + Feedback(Row level) \end{tabular} & 0.8698 & 0.9201 \\ \hline
            \begin{tabular}{@{}l@{}} Llama-3-70b-Ins Zero-shot \\ + Intermediate sub-tasks \\ + Feedback(Cell level) \end{tabular} & \underline{0.9079} & 0.9755 \\ \hline
            \begin{tabular}{@{}l@{}} Llama-3-70b-Ins Zero-shot \\ + Intermediate sub-tasks \\ + Feedback(Cell level) 2-iter\end{tabular} & \textbf{0.9166} & \underline{0.9835} \\ \hline
        \end{tabular}
    \caption{Results on the \texttt{RotoWire} dataset. We bold the best results and underline the second-best.} 
    \label{table:rotowire-our-results}
\end{table}

Table \ref{table:livesum-own} presents the results on the LiveSum dataset where lower values denote better performance for the selected metrics: Root Mean Square Error(RMSE) and Error Rate(ER). The top 4 rows (denoted by $^*$ symbol) contain prior-art results\cite{texttupletable} with an open-license LLM(Llama-2-70B-Chat) and a proprietary LLM(GPT-4) for reference. The rows that follow present our results.

\begin{table*}[!ht]
    \centering
    \fontsize{9}{10}\selectfont 
        \begin{tabular}{l|cccccc}
        \hline 
            \textbf{Model/Strategy} & \multicolumn{2}{c}{\textbf{Easy}} & \multicolumn{2}{c}{\textbf{Medium}} & \multicolumn{2}{c}{\textbf{Hard}} \\
            \cline{2-7}
            & \textbf{RMSE} & \textbf{ER} & \textbf{RMSE} & \textbf{ER} & \textbf{RMSE} & \textbf{ER} \\ \hline
            
            Llama-2-70B-Chat Zero-shot\textsuperscript{*}  & 0.41 & 12.34 & 3.189 & 88.59 & 4.941 & 92.41 \\ \hline
            Llama-2-70B-Chat Zero-shot COT\textsuperscript{*}  & 0.45 & 12.86 & 3.221 & 89.25 & 5.314 & 94.24 \\ \hline
            GPT-4 Zero-shot \textsuperscript{*}  & 0.156 & 4.64 & \textbf{1.167} & 46.05 & 4.114 & 88.53 \\ \hline
            GPT-4 Zero-shot COT\textsuperscript{*}  & 0.154 & 4.38 & \underline{1.173} & \textbf{45.86} & 3.981 & 88.73 \\ \hline \hline
            
            Llama-3-70B-Ins Zero-shot  & 0.312 & 2.91 & 1.751 & 54.67 & 3.568 & 85.97 \\ \hline
            Llama-3-70B-Ins Zero-shot CoT  & 0.229 & 2.58 & 2.216 & 55.42 & 6.364 & 87.99 \\ \hline
            \begin{tabular}{@{}l@{}} Llama-3-70B-Ins Zero-shot \\ + Intermediate Steps \end{tabular}  & 0.084 & 0.76 & 1.549 & 53.73 & 3.374 & 84.84 \\ \hline
            
            \begin{tabular}{@{}l@{}} Llama-3-70B-Ins Zero-shot \\ + Intermediate Steps \\ + Feedback(Cell level) \end{tabular}  & \underline{0.062} & \underline{0.39} & 1.363 & 47.33 & \textbf{2.385} & \underline{63.32} \\ \hline
            \begin{tabular}{@{}l@{}} Llama-3-70B-Ins Zero-shot \\ + Intermediate Steps \\ + Feedback(Cell level) 2-iterations \end{tabular}  & \textbf{0.057} & \textbf{0.31} & 1.290 & \underline{45.98} & \underline{2.504} & \textbf{59.95} \\ \hline 
        \end{tabular}
    \caption{Results on the \texttt{LiveSum} dataset. We bold the best results and underline the second-best. }
    \label{table:livesum-own}
\end{table*}

\noindent \textbf{Intermediate sub-tasks guidance enhances LLM performance in text-to-table generation.}
Our results on both datasets demonstrate that incorporating custom intermediate sub-tasks enhances the reasoning capabilities of the LLM, leading to better performance compared to regular and Chain-of-Thought (CoT) prompting for the text-to-table generation task. The use of explicit intermediate steps tailored for the task offers a clear advantage. This approach not only elevates the performance of the open-license LLM considered, but also brings it on par with larger, proprietary models on text-to-table generation task.

\noindent \textbf{LLM self-feedback on the generated table as a whole is error-prone and less useful.}
Self-refinement using table level feedback showed limited effectiveness. LLM struggles to understand and critique its own table output effectively. The feedback on the entire table or on individual rows often proved either unhelpful or counterproductive. This aligns with findings from \cite{largelanguagemodelsselfcorrect} which suggests that self-improvement guided by LLMs is often infeasible for off-the-shelf models when dealing with complex reasoning tasks. 

\noindent \textbf{Fine-grained cell level feedback improves table accuracy.}
On the other hand, more granular cell-level feedback enabled more accurate corrections. 
LLMs at times overlook incorrect table values initially. However, prompting the model to specifically reassess one cell value at a time often resulted in more accurate values. This is more evident with LiveSum dataset, as it requires complex calculation operations like counting of events to populate the cell values.
This step in a way mirrors the cell-wise QA based prompting approach used by \cite{gtbls} to construct tables accurately, but through a cell level re-verification of values. 

\noindent \textbf{Iterative cell level feedback provides slight performance gain with potential risks.}
Iterative feedback offers slight performance improvement in general. But it can at times introduce noise and complicate the generation process. In our experiments, we stopped the iteration process after two cycles based on these observations. It underscores the need to determine an optimal stopping point in the iteration process to balance improved accuracy with the risk of over-complicating or introducing errors. 

\noindent \textbf{Cost-Performance trade-off.}
On the cost front, row-level feedback makes LLM calls proportional to the number of rows, and cell-level feedback necessitates calls proportional to the number of cells for a single iteration. 
Therefore, despite its potential for more accurate results, self-improvement via cell-level feedback could be less practical for some applications due to the high volume of LLM invocations. In such cases, the initial tables generated through our intermediate sub-task guidance strategy can suffice. However, for applications where high accuracy is crucial, further refinement via cell-level feedback is recommended.

\subsection{Deployment Considerations}
Our system leverages open LLMs via an API interface, avoiding costly fine-tuning and minimizing setup complexity for rapid deployment with minimal resources. Internally, we are focused on deploying it in relevant customer use cases, evaluating performance, conducting user studies, and refining the system based on real-world feedback.

\section{Conclusion \& Future Work}
We presented a system for LLM-driven text-to-table generation that employs custom prompting strategies, which involve guiding the LLM with intermediate sub-tasks and refining the output through iterative self-feedback. We demonstrated that this structured guidance significantly improves model performance for text-to-table generation, compared to standard prompting strategies, while eliminating the need for costly fine-tuning. Additionally, we implemented iterative table output refinement through self-feedback to further enhance LLM performance in text-to-table tasks.
Our system achieves state-of-the-art performance on two complex text-to-table benchmark datasets from the public domain.
We also address the key challenges in LLM self-feedback based output refinement and outline the trade-offs between the cost of LLM invocations and the quality of the generated tables.
Overall, our approach emphasizes the importance of tailored prompting and refined feedback mechanisms for automated text-to-table generation using LLMs. Future research can further enhance these strategies and extend their application to other structured outputs, expanding their utility across diverse domains.


\clearpage
\appendix
\section{Appendix}
\label{sec:appendix}
\subsection{Dataset Details}
Table \ref{table:dataset-stats} provides a high level overview of the dataset statistics. Here `$\#$Pts' refers to number of data points. Table \ref{table:table-stats} provides statistics of tables in both Rotowire \& LiveSum datasets.

\begin{table}[!ht]
    \centering
        \begin{tabular}{ccccc}
        \hline
            \begin{tabular}{@{}c@{}} Dataset \end{tabular} & 
            \begin{tabular}{@{}c@{}}$\#$Pts \\ (Train) \end{tabular} & \begin{tabular}{@{}c@{}}$\#$Pts \\ (Val) \end{tabular} & \begin{tabular}{@{}c@{}}$\#$Pts \\ (Test) \end{tabular} & \begin{tabular}{@{}c@{}}\#Tokens \\ (Avg) \end{tabular} \\ \hline
            Rotowire & 3.4k & 727 & 728 & 351.05 \\ \hline
            LiveSum & 3017 & - & 754 & 1256 \\ \hline
        \end{tabular}
    \caption{Dataset statistics}
    \label{table:dataset-stats}
\end{table}

\begin{table}[!ht]
    \centering
        \begin{tabular}{cccc}
        \hline
            \begin{tabular}{@{}c@{}}Dataset/Table \\ Name \end{tabular} & \begin{tabular}{@{}c@{}}\#Rows \\ (Avg) \end{tabular} & \begin{tabular}{@{}c@{}}\#Cols \\ (Avg) \end{tabular} & \begin{tabular}{@{}c@{}}\#Non-Empty \\ Cells(Avg) \end{tabular} \\ \hline
            RotoWire/Player & 7.26 & 8.75 & 22.63 (43.93\%) \\ 
            RotoWire/Team & 2.71 & 4.84 & 6.56 (85.4\%)\\
            LiveSum & 2 & 8 & 16 \\ \hline
        \end{tabular}
    \caption{Table statistics within Rotowire \& Livesum}
    \label{table:table-stats}
\end{table}

\begin{figure*}[!htbp]
\centering
\includegraphics[width=\linewidth]{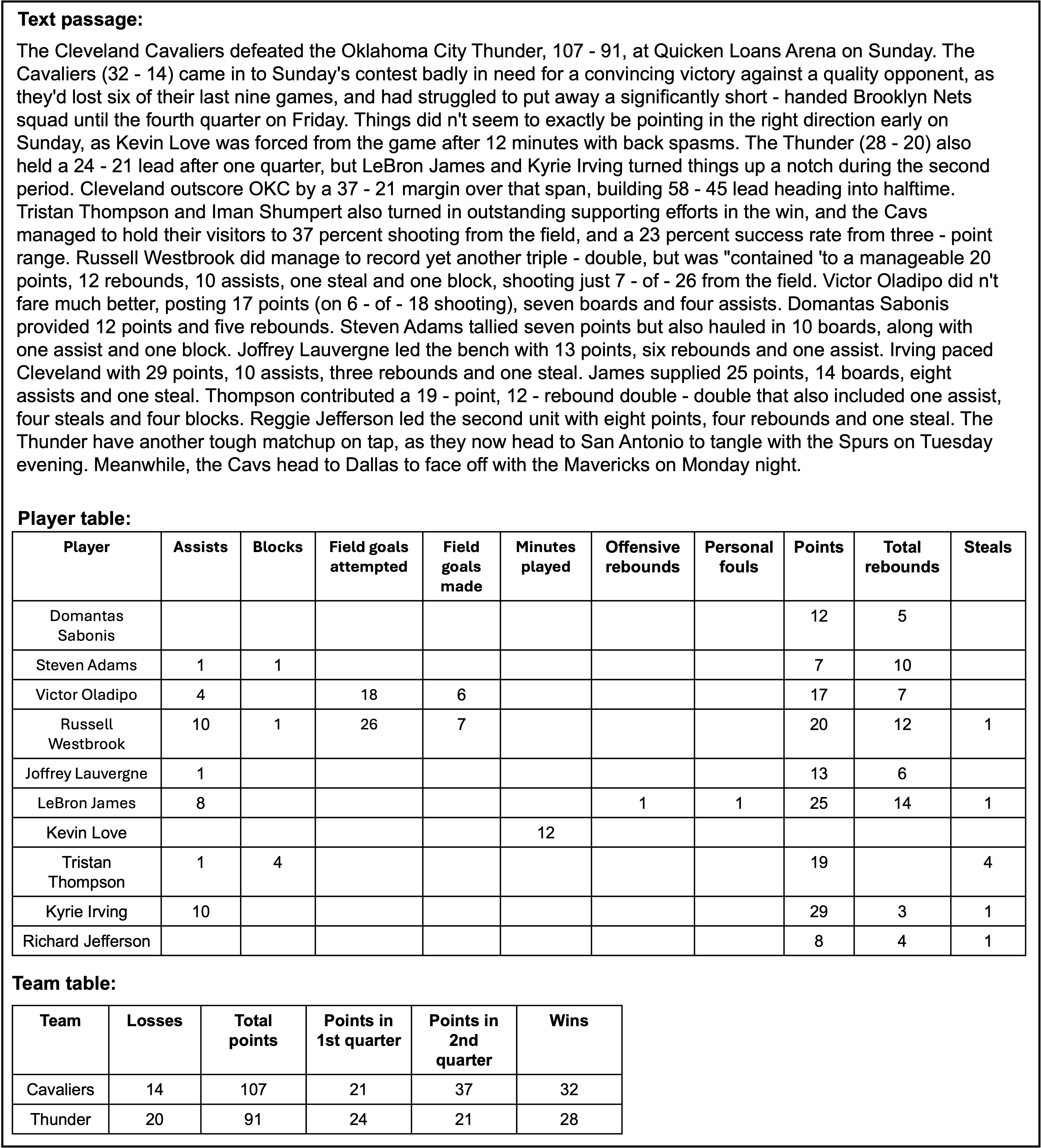}
\caption{An example of a text passage and its corresponding tables from the Rotowire dataset.}
\label{fig:rotowire-example-big}
\end{figure*}

Figure \ref{fig:rotowire-example-big} shows an example from the Rotowire dataset where the player table has 11 columns and 11 rows while the team table has 6 columns and 3 rows including the header.

\begin{figure*}[!htbp]
\centering
\includegraphics[width=\linewidth]{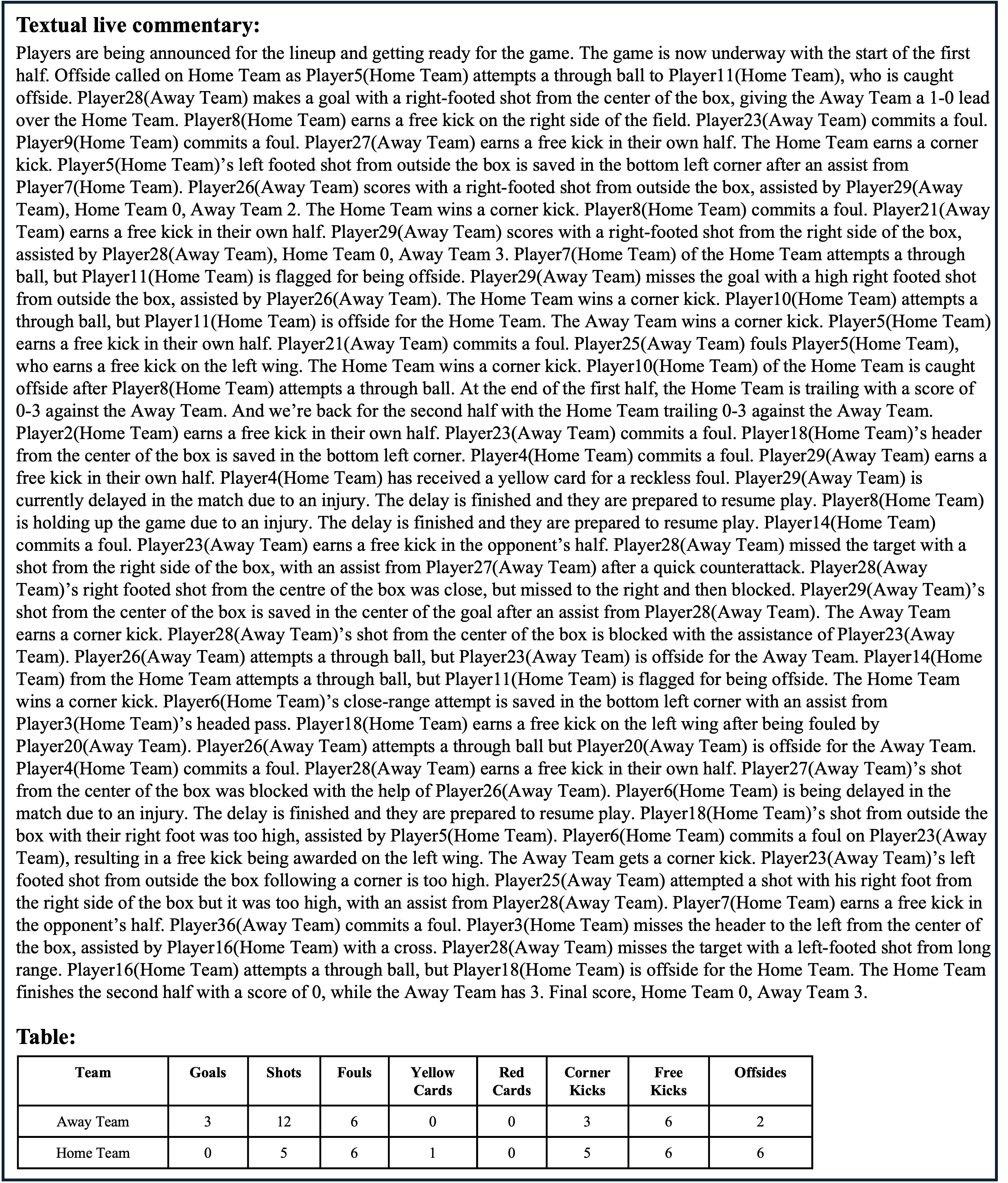}
\caption{An example of a text passage and its corresponding table from the Livesum dataset.}
\label{fig:livesum-example}
\end{figure*}

Figure \ref{fig:livesum-example} shows an example from the LiveSum dataset. Here the table has 9 columns and 3 rows including the header.

\subsection{Prompt Examples}
\begin{figure*}[!htbp]
\centering
\includegraphics[width=\linewidth]{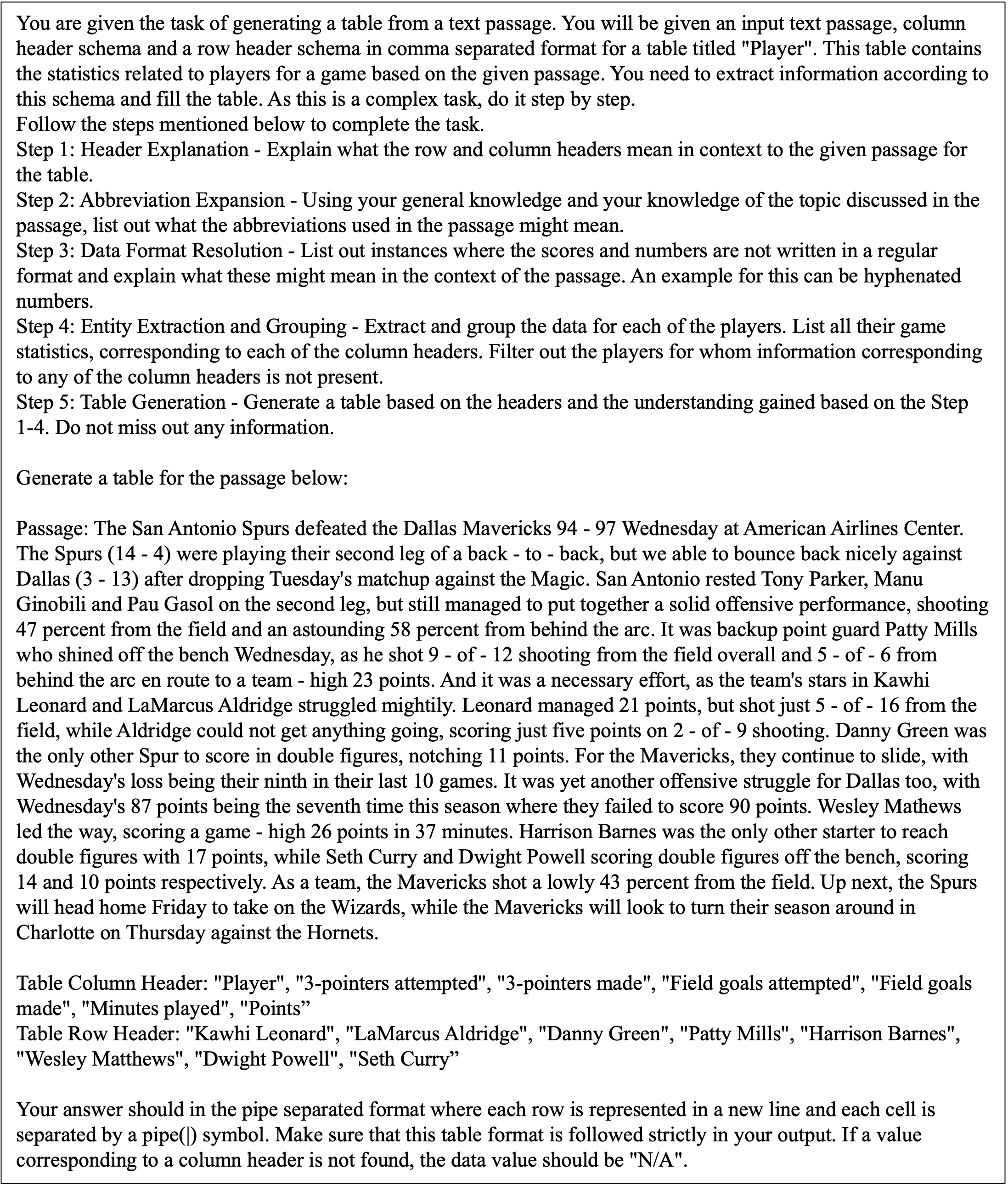}
\caption{An example of a table generation prompt with intermediate subtasks for the Rotowire dataset.}
\label{fig:prompt-example-1}
\end{figure*}

Figure \ref{fig:prompt-example-1} shows an example prompt used for text-to-table generation using intermediate sub-tasks guidance strategy with the Rotowire dataset.

\begin{figure*}[!htbp]
\centering
\includegraphics[width=\linewidth]{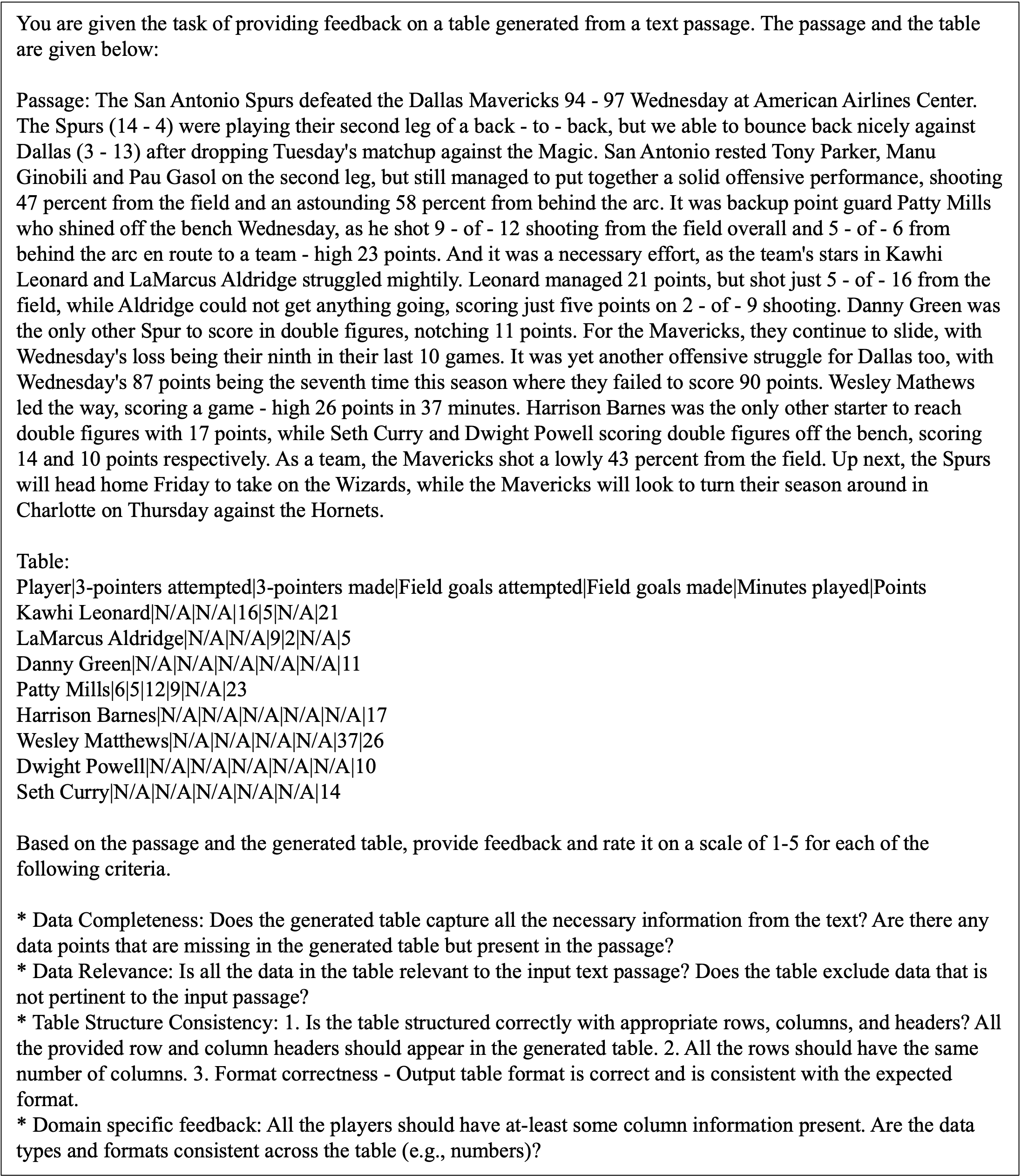}
\caption{An example of a table-level feedback generation prompt with the Rotowire dataset.}
\label{fig:prompt-example-2}
\end{figure*}

Figure \ref{fig:prompt-example-2} shows an example prompt used for getting table-level feedback on the generated table.

\subsection{Model Choices}
In addition to Llama-3-70B-Instruct, one of the most popular open-license LLMs, we also experimented with the Mixtral-8x7b-Instruct model, another open-license LLM via our organization's AI Models framework to assess generalizability. 
Our strategy demonstrated similar relative performance gains with Mixtral-8x7b-Instruct compared to standard prompting approaches. However, frequent inconsistencies in the model's adherence to task instructions limited the reliability of its results for reporting.

\subsection{Limitations and Other Considerations}
We evaluated our system on public datasets from the sports domain, as these are widely used benchmarks in prior text-to-table generation research. Although our approach is not inherently limited to any particular domain, further exploration with datasets from other domains could help assess its adaptability across domains.
Our methodology assumes the presence of both row and column headers, which is particularly suited for entity-centric tables, where rows correspond to specific entities and columns provide their attributes. In scenarios where this assumption does not hold, additional strategies may be required to accurately interpret and structure the input data.
The instruction-following capability of LLMs is crucial for the proposed approaches. While selecting models with strong instruction-following abilities is important, this does not pose a significant limitation on the broader applicability of our system.

\end{document}